\pdfoutput=1

\documentclass[11pt]{article}

\usepackage[preprint]{acl}
\usepackage[T1]{fontenc}
\usepackage[utf8]{inputenc}

\usepackage{amsthm}
\usepackage{amsmath}
\newtheorem{lemma}{Lemma}
\newtheorem{theorem}{Theorem}

\DeclareMathOperator*{\argmin}{arg\,min}

\usepackage{bm}
\usepackage{times}
\usepackage{pifont}
\usepackage{latexsym}
\usepackage{amsfonts}
\usepackage{enumitem}
\usepackage{booktabs}
\usepackage{multirow}
\usepackage{microtype}
\usepackage{inconsolata}

\usepackage{pgfplots}
\usepackage{graphicx}
\usepgfplotslibrary{fillbetween}
\pgfplotsset{compat=1.18}

\usepackage{xcolor}
\definecolor{softred}{RGB}{250,100,100}
\definecolor{softgreen}{RGB}{56,118,29}
\definecolor{softblue}{RGB}{100,150,200}
\definecolor{softyellow}{RGB}{230,210,140}
\newcommand{\textyellow}[1]{{\color{softyellow}#1}}
\newcommand{\textred}[1]{{\color{softred}#1}}
\newcommand{\textblue}[1]{{\color{softblue}#1}}
\newcommand{\textgreen}[1]{{\color{softgreen}#1}}

\usepackage{xspace}
\newcommand{\ourmethod}{\textsc{Format-Adapter}\xspace}

\title{Format-Adapter: Improving Reasoning Capability of LLMs\\by Adapting Suitable Format}

\author{
    Dingzirui Wang$^1$\thanks{Correspondence to: \texttt{\{dzrwang, car\}@ir.hit.edu.cn}}, Xuanliang Zhang$^1$, Rongyu Cao$^2$, Longxu Dou$^2$ \\
    {\bf Xianzhen Luo$^1$}, {\bf Yingwei Ma$^2$}, {\bf Qingfu Zhu$^1$}, {\bf Wanxiang Che$^{1*}$} \\
    {\bf Binhua Li$^2$}, {\bf Fei Huang$^2$}, {\bf Yongbin Li$^2$} \\
    $^1$Harbin Institute of Technology\quad$^2$Independent Researcher \\
}

\begin{document}
    \maketitle
    \begin{abstract}
        Generating and voting multiple answers is an effective method to mitigate reasoning inconsistencies of large language models (LLMs). 
        Prior works have shown that multiple reasoning formats outperform a single format when generating multiple answers. 
        However, previous works using multiple formats rely on formats labeled by humans, which could be unsuitable for all tasks and have high labeling costs. 
        To address this issue, we adapt suitable formats to the given tasks by generating and selecting formats. 
        We first propose how to measure the reasoning error when generating multiple answers. 
        Then, we introduce \ourmethod, which utilizes LLMs to generate and select suitable reasoning formats by minimizing the error measurement we present. 
        We conduct experiments on math and commonsense reasoning tasks, where \ourmethod achieves a $4.3\%$ performance improvement on average over previous works, demonstrating the effectiveness.
    \end{abstract}

    \section{Introduction}
The prior research has revealed that, due to the inconsistency, one question could yield different responses when suffering minor variations in the input or parameters, resulting in incorrect results \cite{wang-etal-2022-measure}. 
To address this issue, previous works propose to generate multiple responses to mitigate the impact of model inconsistencies \cite{wang2023selfconsistency,yao2023tree,besta-etal-2024-graph}.
Specifically, such methods generate multiple answers to a given question by varying parameters and then select the most appropriate response as the final answer by scoring and voting.

\begin{figure}[t]
    \centering
    \includegraphics[width=0.85\linewidth]{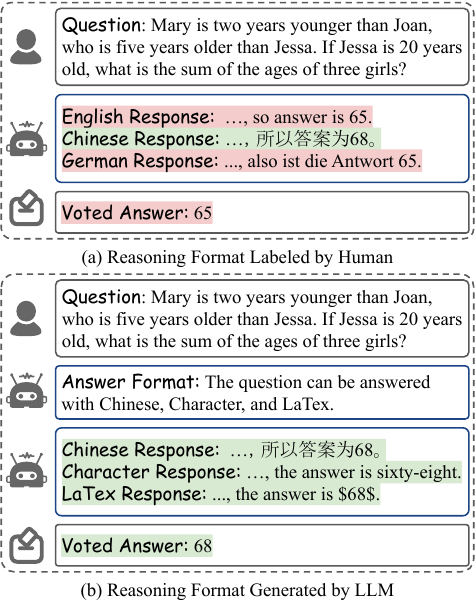}
    \vspace{-0.8em}
    \caption{
        The comparison between the previous work (a) and our method (b) instructed to reason with different formats.
        The \textred{red} parts denote the incorrect answers and the \textgreen{green} parts denote the correct ones.
        The previous work employs the formats labeled by humans, which could be not suitable for the given question and LLM.
        Our method generates and selects suitable formats, achieving better performance.
    }
    \vspace{-1.5em}
    \label{fig:motivation}
\end{figure}

However, the above works rely on the fixed \textit{reasoning format}\footnote{In this paper, we define the \textit{reasoning format} as LLMs how to present the reasoning process.}, which limits the model performance since different questions could suit different reasoning formats \cite{cheng2023binding,chen2023program,he2024doespromptformattingimpact}, as shown in Figure~\ref{fig:motivation}.
Therefore, many prior works try to enhance the reasoning performance by employing various reasoning formats \cite{luo2024pythonbestchoiceembracing,zhang-etal-2024-autocap}.
For example, CLSP~\cite{qin-etal-2023-cross} proposes using varied natural languages to generate different answers in numerical reasoning tasks.
Similarly, FlexTaF~\cite{zhang2024flextafenhancingtablereasoning} addresses table reasoning tasks by generating different answers through diverse table formats.

However, the above methods rely on manually designed reasoning formats, which have the following issues:
\textit{(i)} Manually designed formats could \textbf{not be suitable for the task}; 
\textit{(ii)} Manually designing formats for each task \textbf{incurs significant overhead}.
To address these issues, in this paper: 
\textit{(i)} We discuss \textbf{\textit{why adapting multiple formats outperforms using a single format}} during reasoning;
\textit{(ii)} We propose \textbf{\textit{using LLMs to generate and select suitable formats}} to enhance reasoning performance.

We first propose how to measure the error of the reasoning with multiple responses.
Based on the measurement, we discuss that generating with a single format can only enhance reasoning robustness while using multiple formats can further enhance reasoning capabilities. 
Then, we propose \ourmethod, which utilizes LLMs to generate and select suitable reasoning formats.
We use LLMs to derive reasoning formats without human involvement, lowering the overhead of the format design.
Besides, we propose to adapt reasoning formats by reducing the error measurement we present, ensuring that the format is suitable for the task.

To evaluate the effectiveness of \ourmethod, we adapt our method to two mainstream reasoning tasks: math reasoning (GSM8K~\cite{cobbe2021gsm8k}, MATH~\cite{saxton2018math}) and commonsense reasoning (ARC-Challenge~\cite{yadav-etal-2019-arc}, GPQA~\cite{yadav-etal-2019-arc}).
The experimental results show that, compared with baselines with the single format, \ourmethod brings $4.1\%$ performance improvement on average, proving the effectiveness of \ourmethod.
We also compare \ourmethod with baselines using multiple reasoning formats, where our method brings $4.7\%$ improvement on average, showing the necessity of the format selection.

Our contributions are as follows:
\begin{itemize}[nolistsep,leftmargin=*]
    \item To shed light on further research, we discuss why generating multiple answers with multiple formats outperforms single format;
    \item To enhance the reasoning ability of LLMs, we present \ourmethod, which generates and selects suitable formats using LLMs;
    \item Experiments show that our method brings $4.3\%$ improvement on average over all baselines, showing the effectiveness of \ourmethod.
\end{itemize}

    \section{Preliminaries}
        \label{sec:analysis}

To prove the effectiveness of employing multiple reasoning formats and shed light on future research, in this section, we discuss: 
\textit{(i)} How to measure the error of reasoning with a single reasoning format of LLMs;
\textit{(ii)} How to measure the error of reasoning employing multiple reasoning formats of LLMs and why it outperforms using the single format.

\begin{figure}[t]
    \centering
    \includegraphics[width=0.8\linewidth]{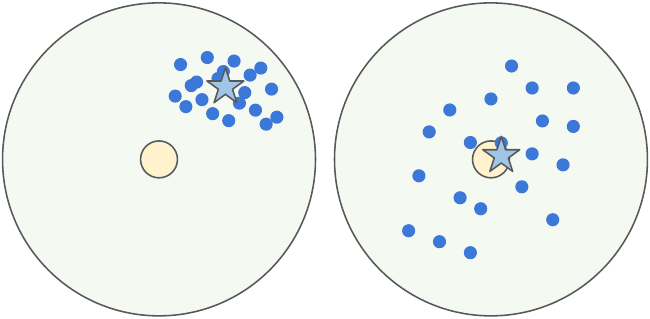}
    \caption{
        The comparison between using the single format (left) and multiple formats (right) with the same number of generated answers.
        The \textyellow{yellow $\bigcirc$} denotes the correct answer, the \textblue{blue $\bullet$} denotes different predictions, and the \textblue{blue \ding{73}} denotes the average prediction.
    }
    \vspace{-1em}
    \label{fig:analysis}
\end{figure}

\subsection{Error of Single Reasoning Format}
    \label{subsec:error_single_format}

    First, we discuss the error of the general ensemble method \cite{sagi_etal_2018_ensemble}, since generating multiple responses and voting can be regarded as an ensemble method.
    In this paper, we use the error function \( L(x, y) \) as follows:
    \begin{equation}
        L(x, y) = \begin{cases} 
            1 & \text{if } x \neq y \\ 
            0 & \text{if } x = y 
        \end{cases}
    \end{equation}
    The function $L$ represents whether the prediction result matches the correct answer exactly.
    We define \( D = \{(x_i, y_i)\}_{|D|} \) as the experimental dataset, \( \{\phi_i\}_m \) as the set of \( m \) predictors, and \( \bar{\phi} = \texttt{avg}({\phi_i}_m) \) as the ensemble predictors.
    Proved by \citet{wood-etal-2024-unified}, the error of ensemble learning with \( m \) predictors on the dataset $D$ can be expressed as the error over the dataset minus the divergence among the individual predictors, that is: 
    \begin{equation}
        \begin{aligned}
            \mathbb{E}_D \left[ L(\bar{\phi}, y) \right] & = \frac{1}{m} \sum_{i=1}^{m} \mathbb{E}_D \left[ L \left( \phi_i, y \right) \right] \\
            & - \mathbb{E}_D \left[ \frac{1}{m} \sum_{i=1}^{m} L \left( \phi_i, \bar{\phi} \right) \right]
        \end{aligned}
        \label{equ:ensemble_error}
    \end{equation}

    Then we discuss the error of generating multiple responses using LLMs.
    For a model employing a single reasoning format, we assume the used format is $\texttt{f}$ and the model is $\phi$. 
    Since only parameters (e.g., random seed, temperature) are altered during reasoning, we can regard the predictor as applying a perturbation to the model inherent performance $\phi \circ \texttt{f}$, expressed as $\phi_i = \phi \circ \texttt{f} + \delta_i$, where $\delta_i$ denotes the perturbation. 
    It can be derived that generating one single answer using $\phi_i$ can be present as:
    \begin{equation}
        \mathbb{E}_D \left[ L(\phi_i, y) \right] = \mathbb{E}_D \left[ L \left( \phi \circ \texttt{f} + \delta_i, y \right) \right]
        \label{equ:single_answer_error}
    \end{equation}

    We assume an ideal scenario where the average of all predictors represents the inherent performance of the model, i.e., $\lim_{m \to \infty} \bar{\phi} = \phi \circ \texttt{f}$.
    It can be proven that the error in generating multiple answers using a single reasoning format satisfies:
    \begin{equation}
        \mathbb{E}_D \left[ L(\bar{\phi}, y) \right] = \mathbb{E}_D \left[ L \left( \phi \circ \texttt{f}, y \right) \right]
        \label{equ:single_format_error}
    \end{equation}

    Appendix~\ref{app:single_error_prove} presents the prove of Equation~\ref{equ:single_format_error}.
    It can be seen that, compared with Equation~\ref{equ:single_answer_error}, generating multiple answers can eliminate the perturbation $\delta$, enhancing the robustness.
    However, when using single format $f$, Equation~\ref{equ:single_format_error} is determined by $\phi$, showing that enhancing performance relies on improving the model capability.

\subsection{Error of Multiple Reasoning Format}
    \label{subsec:error_multiple_format}

    In the following, we discuss the error of using multiple reasoning formats and why it outperforms the single format. 
    During reasoning, we employ multiple formats $\{f_i\}_m$ with one single model $\phi$, so we can assume the predictors to be $\phi_i = \phi \circ \texttt{f}_i + \delta_i$.
    It can be proved that the reasoning error follows:
    \begin{equation}
        \begin{aligned}
            \mathbb{E}_{D} \left[ L(\bar{\phi}, y) \right] & = \frac{1}{m} \sum_{i=1}^m \mathbb{E}_{D} \left[ L(\phi \circ \texttt{f}_i, y) \right] \\
            & - \mathbb{E}_D[\frac{1}{m} \sum_{i=1}^m L(\phi \circ \texttt{f}_i, \bar{\phi})]
        \end{aligned}
        \label{equ:multiple_format_error}
    \end{equation}

    The prove of Equation~\ref{equ:multiple_format_error} can be seen in Appendix~\ref{app:multiple_error_prove}.
    In the equation, the first term denotes the average error over all predictions and correct answers, and the second term measures the divergence between different reasoning formats.
    It can be observed that, even with the same model, we can combine different reasoning formats to minimize error, thereby improving performance, as shown in the right part of Figure~\ref{fig:analysis}.

    \section{Methodology}
        \begin{figure*}[t]
    \centering
    \includegraphics[width=\linewidth]{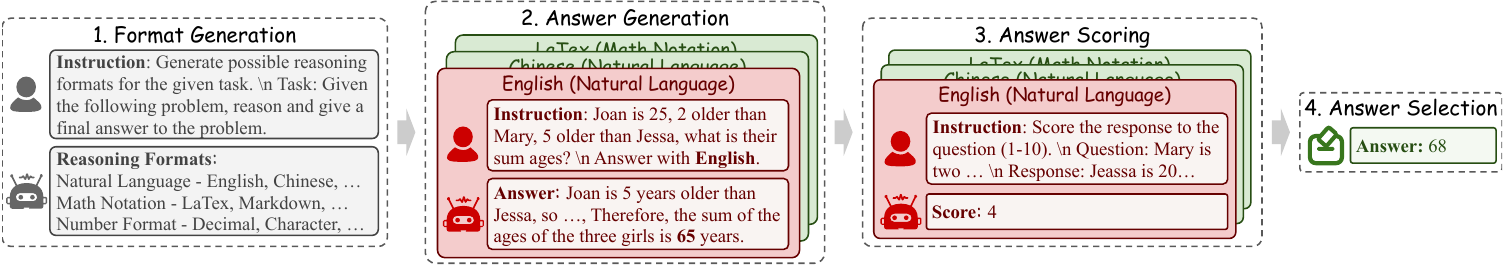}
    \caption{
        The pipeline of \ourmethod, which consists of:
        \textit{(i) Format Generation}: Generate possible reasoning formats of the given task;
        \textit{(ii) Answer Generation}: Generate the answer using each reasoning format;
        \textit{(iii) Answer Scoring}: Score whether each generated answer is correct using LLMs;
        \textit{(iv) Answer Selection}: Select the final answer with Equation~\ref{equ:multiple_format_error}.
        \textred{Red} and \textgreen{green} represent the reasoning formats of incorrect and correct respectively.
    }
    \vspace{-1em}
    \label{fig:method}
\end{figure*}

This section introduces \ourmethod, which leverages LLMs to generate and select the suitable reasoning formats to enhance reasoning performance.
An illustration of our method is shown in Figure~\ref{fig:method}. 
The prompts we used are provided in Appendix~\ref{app:prompt}.
\textbf{\textit{We also discuss the efficiency of \ourmethod in Appendix~\ref{app:efficiency}.}}

\subsection{Format Generation}
    This step is designed to generate reasoning formats, ensuring both the relevancy and diversity of the generated formats.
    Relevancy means that the generated reasoning formats are relevant to the given task. 
    Diversity demands that the generated reasoning formats be varied to ensure suitable reasoning formats for various user questions.

    To ensure relevancy, an example task is provided during generation to help LLMs learn how to produce task-relevant reasoning formats. 
    To ensure diversity, we design the instruction to ask LLMs to generate reasoning formats across multiple categories, where each category consists of multiple formats. 
    For instance, as shown in Figure~\ref{fig:method}, \textit{Natural Language} is the reasoning format category, while \textit{English} and \textit{Chinese} are the reasoning formats of this category.
    In summary, the input includes the task definition, and an example of the task, while the output consists of multiple reasoning formats.
    Appendix~\ref{app:reasoning_format} discusses the generated formats under each setting.

\subsection{Answer Generation}
    This step generates corresponding answers for each generated reasoning format. 
    First, the instruction is rewritten according to each reasoning format to ensure that the answer generation follows the given reasoning format. 
    We take the original instruction of the task (Appendix~\ref{app:prompt}) and the reasoning format as input and ask LLMs to output the rewritten instruction based on the reasoning format.
    Then, the rewritten instruction is used to generate different reasoning answers for the given user question. 
    Following prior work \cite{qin-etal-2023-cross}, zero-shot learning is applied by inputting the rewritten instruction and the user question to output answers in the specified reasoning format.

\subsection{Answer Scoring}
    \label{subsec:method_answer_score}

    After obtaining answers in different reasoning formats, based on Equation~\ref{equ:multiple_format_error}, we aim to select the suitable reasoning formats that minimize the error. 
    However, in Equation~\ref{equ:multiple_format_error}, the error between the prediction and the answer $L(\phi \circ f_i, y)$ is difficult to compute, as the correct answer $y$ is unknown. 
    Therefore, we use answer generation LLMs to score the answers in each reasoning format to estimate the probability that the predicted answer is correct. 
    Following \citet{zheng2023judging}, we input the user question and the predicted answer, outputting a score from $1$ to $10$ to represent the degree to the probability that the answer is correct.
    To ensure that there is the same scale between the first term and the second term of Equation~\ref{equ:multiple_format_error} during the calculation, we divide the rating by $10$ to correspond to the interval of $[0, 1]$.

\subsection{Answer Selection}
    Based on the predicted answers and corresponding scores of different reasoning formats, we discuss how to select the final answers based on Equation~\ref{equ:multiple_format_error}. 
    Specifically, given the dataset $D$ and the model $\phi$, we hope to find suitable reasoning formats to minimize the error that satisfies:

    \begin{equation}
        \{\texttt{f}_i\}_n = \argmin_{\{\texttt{f}_i\}_n \subseteq \{\texttt{f}_i\}_m} \mathbb{E}_{D} \left[ L(\bar{q}, y) \right]
        \label{equ:answer_selection}
    \end{equation}
    
    In Equation~\ref{equ:multiple_format_error}, the first term can be directly calculated by averaging scores obtained in \S\ref{subsec:method_answer_score}.
    The second term requires calculating the average difference between all results and the average result, where we take the average prediction $\bar{\phi}(x)$ as the answer appearing most frequently among all outcomes.
    Considering computational efficiency, we adopt a greedy algorithm to select formats: we add each format \( \texttt{f}_i \) one by one to the selected results, where if the value of Equation~\ref{equ:multiple_format_error} decreases, we retain \( \texttt{f}_i \); otherwise, we remove \( \texttt{f}_i \).
    Due to the inherent scoring errors of LLMs, we do not directly use the answer with the highest score within the selected set. 
    Instead, we choose the most frequently occurring answer as the final answer.

    \section{Experiment}
        \subsection{Experimental Setup}
    \subsubsection{Datasets}
        To validate the effectiveness of our method, following \citet{dubey2024llama3herdmodels}, we conduct experiments on two mainstream reasoning tasks: commonsense reasoning (ARC-Challenge-Hard~\cite{yadav-etal-2019-arc}, GPQA~\cite{rein2024gpqa}) and math reasoning (GSM8K~\cite{cobbe2021gsm8k}, MATH~\cite{saxton2018math}).
        Commonsense reasoning requires the model to apply commonsense knowledge to comprehend and answer questions.
        On the other hand, math reasoning demands the model to solve mathematical problems.

        Due to the high cost of generating multiple answers, we employ the subsets of the above benchmarks to reduce computational overhead while maintaining a robust performance evaluation. 
        Specifically, for GSM8K and ARC-Challenge (ARC-C), we sample $256$ questions that are not well solved by the current LLMs, referred to as GSM8K-Hard and ARC-C-Hard, respectively.
        For MATH, we utilize the version of MATH500~\cite{lightman2024lets}, which samples $500$ questions from the original dataset.
        For GPQA, we employ the original test set, comprising $448$ questions.

    \subsubsection{Models}
        We conduct the experiments with the models of Llama3.1-Instruct~\cite{dubey2024llama3herdmodels} and GPT-4o~\cite{openai2024gpt4technicalreport}. 
        Llama3.1 is one of the most mainstream and high-performing open-source LLMs. 
        GPT-4o, on the other hand, currently represents one of the most powerful LLMs in terms of reasoning capabilities. 
        Our selection ensures coverage of diverse application scenarios.

    \subsubsection{Baselines}
        To better reflect the effectiveness of \ourmethod, we compare our method with two types of baselines.
        The first type uses the single reasoning format, including Single, Self-Consistency (SC) \cite{wang2023selfconsistency}, Tree-of-Thought (ToT) \cite{yao2023tree}, and DTV \cite{zhou2024dtv}.
        Another type uses multiple reasoning formats, including CLSP~\cite{qin-etal-2023-cross}, MultiPoT~\cite{luo2024pythonbestchoiceembracing}, and FlexTaF~\cite{zhang2024flextafenhancingtablereasoning}.
        We introduce the above baselines in Appendix~\ref{app:baseline}. 

    \subsubsection{Metrics}
        We use Exact Match (EM) as the evaluation metric for all datasets, which measures whether the predicted result is completely identical to the ground truth. 
        Additionally, we evaluate methods that generate multiple answers under two settings: Vote and Oracle. 
        Vote refers to selecting one answer from all generated answers as the final result, reflecting the actual performance of the method. 
        Oracle, on the other hand, considers a question correct if there exists one of the generated answers matches the ground truth, reflecting the performance upper bound of the method.

    \subsubsection{Implement Details}
        Following the previous work \cite{qin-etal-2023-cross}, we evaluate \ourmethod using zero-shot.
        The numbers and types of reasoning formats of \ourmethod under different settings can be seen in Appendix~\ref{app:reasoning_format}.
        The parameter settings of \ourmethod are consistent with Single to ensure comparable results.
        To verify the robustness of \ourmethod, we show the average result with five running in Appendix~\ref{app:robustness}.

\subsection{Main Experiment}
    \subsubsection{Baselines with Single Format}
        \begin{table*}[htp]
            \centering
            \small
            \begin{tabular}{ll|cc|cc|cc|cc}
    \toprule
    \multirow{2}{*}{\textbf{Model}} & \multirow{2}{*}{\textbf{Method}} & \multicolumn{2}{c}{\textbf{GSM8K-Hard}} & \multicolumn{2}{|c}{\textbf{MATH500}} & \multicolumn{2}{|c}{\textbf{ARC-C-Hard}} & \multicolumn{2}{|c}{\textbf{GPQA}} \\
     &  & \textbf{Vote} & \textbf{Oracle} & \textbf{Vote} & \textbf{Oracle} & \textbf{Vote} & \textbf{Oracle} & \textbf{Vote} & \textbf{Oracle} \\
    \midrule
    \multirow{5}{*}{Llama3.1-8b} & Single & $23.0$ & $-$ & $47.8$ & $-$ & $16.8$ & $-$ & $28.1$ & $-$ \\
     & SC & $36.7$ & $55.1$ & $51.4$ & $63.0$ & $18.4$ & $23.4$ & $30.8$ & $59.2$ \\
     & ToT & $43.0$ & $53.9$ & $52.8$ & $56.8$ & $24.2$ & $33.8$ & $32.8$ & $46.7$ \\
     & DTV & $51.6$ & $56.2$ & $55.4$ & $56.4$ & $39.1$ & $42.6$ & $32.6$ & $50.0$ \\
     & \ourmethod & $\bm{54.7}$ & $\bm{89.8}$ & $\bm{56.8}$ & $\bm{75.0}$ & $\bm{57.4}$ & $\bm{91.4}$ & $\bm{33.9}$ & $\bm{93.8}$ \\
    \midrule
    \multirow{5}{*}{Llama3.1-70b} & Single & $66.0$ & $-$ & $63.4$ & $-$ & $68.0$ & $-$ & $43.1$ & $-$ \\
     & SC & $70.3$ & $77.3$ & $64.4$ & $72.8$ & $69.1$ & $69.9$ & $46.2$ & $66.5$ \\
     & ToT & $71.5$ & $77.6$ & $67.2$ & $75.2$ & $70.7$ & $72.3$ & $48.0$ & $73.2$ \\
     & DTV & $71.7$ & $84.3$ & $65.8$ & $81.8$ & $69.9$ & $73.8$ & $50.2$ & $75.9$ \\
     & \ourmethod & $\bm{76.2}$ & $\bm{94.9}$ & $\bm{70.4}$ & $\bm{85.4}$ & $\bm{71.5}$ & $\bm{88.7}$ & $\bm{51.0}$ & $\bm{96.4}$ \\
    \midrule
    \multirow{3}{*}{GPT-4o} & Single & $73.4$ & $-$ & $71.0$ & $-$ & $77.0$ & $-$ & $48.9$ & $-$ \\
     & SC & $74.1$ & $82.8$ & $71.4$ & $83.2$ & $78.9$ & $83.2$ & $49.1$ & $70.8$ \\
     & \ourmethod & $\bm{78.4}$ & $\bm{95.1}$ & \bm{$76.8$} & \bm{$86.6$} & \bm{$80.1$} & \bm{$96.9$} & \bm{$51.6$} & \bm{$96.6$} \\
    \bottomrule
\end{tabular}

            \caption{
                EM of \ourmethod and baselines using the single reasoning format.
                The best results of each setting are marked in \textbf{bold}.
                Due to the limitations of computing resources, we only compare the performance of \ourmethod with Self-Consistency on GPT-4o.
            }
            \vspace{-1.5em}
            \label{tab:main_experiment_single}
        \end{table*}

        The experimental results of \ourmethod compared with baselines using the single reasoning format (Appendix~\ref{app:baseline}) are shown in Table~\ref{tab:main_experiment_single}.
        The table shows that, compared with the best baseline results under each setting, \ourmethod brings $4.1\%$ performance improvement on average, showing the effectiveness of \ourmethod.
        We also compare the efficiency across different methods in Appendix~\ref{app:efficiency_output_tokens}.
        Besides, from Table~\ref{tab:main_experiment_single}, we can also observe that:

        \paragraph{Model}
            The improvement brought by \ourmethod on different models depends on the difficulty of the dataset.
            For relatively simple datasets like GSM8K and ARC-Challenge, our method demonstrates more significant improvements on models with a small scale.
            Conversely, for more challenging datasets such as MATH and GPQA, our method achieves more notable improvements on larger models. 
            This is because, for complex datasets, smaller models lack the necessary knowledge to solve such problems due to their limited scale, where simply altering the reasoning format cannot introduce new knowledge, leading to negligible performance gains. 
            On the other hand, models with larger scales already exhibit strong performance for simpler datasets, making the improvements brought by our method less pronounced compared to smaller models.

        \paragraph{Metric}
            \ourmethod demonstrates performance improvements in both the Vote and Oracle settings, indicating that our method not only enhances actual performance but also effectively encourages the model to utilize diverse reasoning formats to generate correct answers. 
            These results also confirm that the most suitable reasoning format varies across different types of questions.
            However, there remains a significant performance gap between the Vote and Oracle settings in our method, which can be attributed to the following reasons: 
            \textit{(i)} The scoring quality of LLMs is suboptimal, making it challenging to accurately assess whether a predicted result is correct;
            \textit{(ii)} Specifically, for datasets such as ARC-Challenge and GPQA, where the answers are choices, LLMs could produce correct results while following incorrect reasoning processes, resulting in lower scores, which can also explain why the performance gap between Vote and Oracle is larger on the commonsense reasoning datasets compared to math reasoning datasets.

    \subsubsection{Baselines with Multiple Format}
        \begin{table}[t]
            \centering
            \small
            \begin{tabular}{ll|cc|cc}
    \toprule
    \multirow{2}{*}{\textbf{Dataset}} & \multirow{2}{*}{\textbf{Method}} & \multicolumn{2}{c}{\textbf{8b}} & \multicolumn{2}{c}{\textbf{70b}} \\
     & & \textbf{Vote} & \textbf{Oracle} & \textbf{Vote} & \textbf{Oracle} \\
    \midrule
    \multirow{3}{*}{MATH} & CLSP & $53.0$ & $66.2$ & $66.9$ & $74.9$ \\
     & MultiPoT & $57.4$ & $66.5$ & $72.2$ & $79.1$ \\
     & Ours & \bm{$60.1$} & \bm{$88.6$} & \bm{$77.2$} & \bm{$88.6$} \\
    \midrule
    \multirow{2}{*}{WikiTQ} & FlexTaF & $38.0$ & $70.3$ & $41.5$ & $79.0$ \\
     & Ours & \bm{$55.2$} & \bm{$79.7$} & \bm{$63.1$} & \bm{$84.0$} \\
    \bottomrule
\end{tabular}

            \caption{
                EM of \ourmethod (Ours) and baselines using multiple reasoning formats on Llama3.1.
                The best results of each setting are marked in \textbf{bold}.
            }
            \vspace{-1em}
            \label{tab:main_experiment_multiple}
        \end{table}

        The performance comparison between \ourmethod and baselines employing multiple reasoning formats is presented in Table~\ref{tab:main_experiment_multiple}.
        Following the setup of MultiPoT, we select $263$ problems from MATH500 that can be resolved using code-based solutions.
        About FlexTaF, we conduct experiments on the WikiTQ dataset~\cite{pasupat-liang-2015-wikitq}, which is the mainstream benchmark of the table QA task.
        As observed from the table, \ourmethod achieves an average improvement of $4.7\%$ over the best performance of the baselines under each setting, demonstrating the effectiveness of our method.

\subsection{Ablation Study}
    \begin{table*}[t]
        \centering
        \small
        \begin{tabular}{ll|llll}
    \toprule
    \textbf{Model} & \textbf{Method} & \textbf{GSM8K-Hard} & \textbf{MATH500} & \textbf{ARC-C-Hard} & \textbf{GPQA} \\
    \midrule
    \multirow{4}{*}{Llama3.1-8b} & \ourmethod & $54.7$ & $56.8$ & $57.4$ & $33.5$ \\
     & \quad - Rewrite & $51.6 \, (-3.1)$ & $53.6 \, (-3.2)$ & $52.0 \, (-5.4)$ & $32.4 \, (-1.1)$ \\
     & \quad - Select & $49.2 \, (-5.5)$ & $47.8 \, (-9.0)$ & $54.7 \, (-2.7)$ & $25.4 \, (-8.1)$ \\
     & \quad - Score & $53.9 \, (-0.8)$ & $54.2 \, (-2.6)$ & $52.7 \, (-4.7)$ & $30.1 \, (-3.4)$ \\
    \midrule
    \multirow{4}{*}{Llama3.1-70b} & \ourmethod & $76.2$ & $70.4$ & $71.5$ & $51.0$ \\
      & \quad - Rewrite & $75.4 \, (-0.8)$ & $69.6 \, (-0.8)$ & $68.8 \, (-2.7)$ & $47.1 \, (-3.9)$ \\
      & \quad - Select & $75.0 \, (-1.2)$ & $68.6 \, (-1.8)$ & $66.4 \, (-5.1)$ & $44.2 \, (-6.8)$ \\
      & \quad - Score & $73.8 \, (-2.4)$ & $70.2 \, (-0.2)$ & $69.9 \, (-1.6)$ & $47.3 \, (-3.7)$ \\
    \bottomrule
\end{tabular}

        \caption{
            The ablation study results under:
            \textit{(i)} Rewrite: Generate answers without rewriting instructions;
            \textit{(ii)} Select: Vote the answer from the responses with the highest score;
            \textit{(iii)} Score: Set all answers with the same score of $1.0$.
        }
        \label{tab:ablation_study}
    \end{table*}

    To demonstrate the effectiveness of each step of \ourmethod, we conduct ablation studies to verify that \ourmethod is the most effective under different answer selection strategies.
    The experimental results are shown in Table~\ref{tab:ablation_study}.
    From the table, we can observe that removing any individual step results in a performance decline, thereby validating the importance of each step in \ourmethod. 
    Furthermore, the table reveals the following insights:
    \textit{(i)} Removing the Select step leads to the most significant performance drop, indicating that in many questions, only a few reasoning formats yield correct answers, necessitating the Select step to identify the correct solutions;
    \textit{(ii)} The performance degradation of the ablation study is more pronounced in smaller-scale models compared to larger ones, which suggests that models with smaller scales generate fewer reasoning formats capable of producing correct answers, relying more heavily on the Rewrite and Select steps to achieve correct results.

\subsection{Analysis}
    \label{subsec:analysis_experiment}

    In this section, we adapt analysis experiments to understand better how \ourmethod improves the reasoning performance and to guide the parameter selection.
    Due to the high reasoning cost, we only employ Llama3.1 as the experimental LLMs.
    We also adapt the case study to understand better how \ourmethod improves the performance, which is discussed in Appendix~\ref{app:case_study}.

    \subsubsection{Reasoning Error}
        \begin{figure}
            \centering
            \small
            \begin{tikzpicture}
    \small
    \begin{axis}[
        xlabel={Value of Equation~\ref{equ:multiple_format_error}},
        ylabel={EM of MATH500},
        ymin=40, ymax=46,
        xmin=0.10, xmax=0.26,
        xtick={0.10, 0.14, 0.18, 0.22, 0.26},
        legend style={at={(0.5,-0.2)},anchor=north,legend columns=2},
        ymajorgrids=true,
        grid style=dashed,
        width=0.8\linewidth,
        every axis legend/.append style={nodes={anchor=west}},
        tick scale binop=\times,
        scaled ticks=false,
        xticklabel style={
            /pgf/number format/fixed,
            /pgf/number format/precision=3
        },
        yticklabel style={
            /pgf/number format/fixed,
            /pgf/number format/precision=1
        }
    ]

    \addplot[only marks, color=softblue, mark size=2pt, mark=*] coordinates {
        (0.110, 45.2)
        (0.124, 43.6)
        (0.163, 43.0)
        (0.179, 42.2)
        (0.188, 42.2)
        (0.201, 44.2)
        (0.213, 42.8)
        (0.214, 42.4)
        (0.219, 41.8)
        (0.219, 42.8)
        (0.224, 43.6)
        (0.226, 43.8)
        (0.228, 43.0)
        (0.230, 42.2)
        (0.236, 43.4)
        (0.238, 41.8)
        (0.240, 42.4)
        (0.241, 40.6)
        (0.247, 40.4)
        (0.252, 41.8)
    };

    \addplot [domain=0.1:0.26, samples=100, color=gray, thick] {46.60 - 18.56*x};

    \addplot [name path=upper, domain=0.1:0.26, samples=100, color=gray, dashed] {48.20 - 18.56*x};
    \addplot [name path=lower, domain=0.1:0.26, samples=100, color=gray, dashed] {44.99 - 18.56*x};

    \addplot [gray!10] fill between[of=upper and lower];

    \end{axis}
\end{tikzpicture}
            \caption{
                The performance on MATH with different formats using Llama3.1-8b.
                Different \textblue{blue $\bullet$} denotes the result using different formats, where the formats used are randomly sampled from that generated by \ourmethod.
                The correlation coefficient is $-0.652$.
            }
            \vspace{-1em}
            \label{fig:reasoning_error}
        \end{figure}
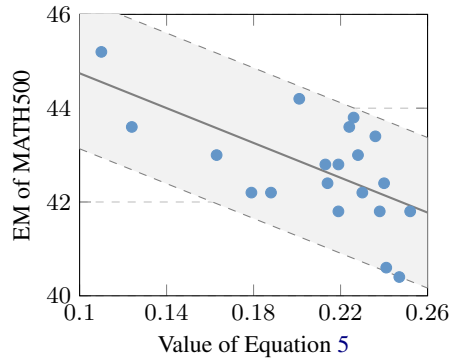
    
        To demonstrate that Equation~\ref{equ:multiple_format_error} effectively reflects the reasoning error, we conduct statistical analysis on MATH to evaluate the model performance corresponding to different values of Equation~\ref{equ:multiple_format_error}.
        The experimental results are shown in Figure~\ref{fig:reasoning_error}, from which we can observe that:
        \textit{(i)} As the value of Equation~\ref{equ:multiple_format_error} gradually increases, the model performance consistently declines, indicating that the error is indeed progressively growing;
        \textit{(ii)} The error obtained in Figure~\ref{fig:reasoning_error} is predominantly concentrated around $0.22$, suggesting that most reasoning formats yield similar results, while these results are inferior to the best results, indicating that the majority of reasoning formats do not produce the correct answers, showing the necessity to select the most suitable reasoning format for each question.

    \subsubsection{Reasoning Format Category}
        \label{subsubsec:reason_format_category}

        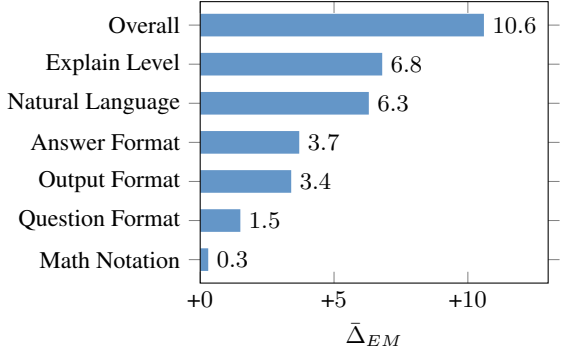
\begin{figure}
            \centering
            \small
            \begin{tikzpicture}
    \begin{axis}[
        xbar,
        xmin=0, xmax=13,
        width=0.8\linewidth,
        xlabel={$\Bar{\Delta}_{EM}$},
        symbolic y coords={Math Notation, Question Format, Output Format, Answer Format, Natural Language, Explain Level, Overall},
        ytick=data,
        nodes near coords,
        nodes near coords align={horizontal},
        bar width=0.3cm,
        axis on top,
        xticklabel={+\pgfmathprintnumber{\tick}}
        ]
        \addplot[draw=none, fill=softblue] coordinates {
            (10.6,Overall)
            (0.3,Math Notation)
            (1.5,Question Format)
            (3.4,Output Format)
            (3.7,Answer Format)
            (6.3,Natural Language)
            (6.8,Explain Level)
        };
    \end{axis}
\end{tikzpicture}
            \vspace{-1em}
            \caption{
                The average improvement brought by \ourmethod with different reasoning categories having more than four formats.
                $\bar{\Delta}_{EM}$ denotes the average improvement compared to Self-Consistency.
                Overall denotes using all reasoning categories.
            }
            \vspace{-1em}
            \label{fig:format_category}
        \end{figure}

        To examine the performance of different reasoning formats and inspire future work, we analyze the average performance improvement achieved under various settings with different reasoning format categories.
        We also list the most suitable reasoning format for each task in Appendix~\ref{app:reasoning_format}.
        The results are shown in Figure~\ref{fig:format_category}, from which we can see that:
        \textit{(i)} For the results using the single reasoning format category, its performance improvement is determined by the variation in answers generated by the corresponding formats of this category. For the categories with low improvements (e.g., Math Notation), the answers across different formats are largely similar, with performance close to that of Self-Consistency. In contrast, reasoning categories with higher performance improvements (e.g., Explain Level) exhibit greater variability in the answers generated by different formats, making it more likely to include the correct result;
        \textit{(ii)} Even for the best-performing single category, its performance improvement is still lower than that achieved by using all reasoning categories (Overall), which indicates that the most suitable reasoning formats vary across questions, and combining different reasoning categories and formats during reasoning is necessary to achieve optimal results.

    \subsubsection{Reasoning Format Scale}
        \begin{figure}
            \centering
            \small
            \begin{tikzpicture}
    \small
    \begin{axis}[
        axis y line*=left,
        xlabel={Format Scale},
        xmin=1, xmax=25,
        ymin=0, ymax=10,
        ylabel={$\Delta_\texttt{EM}$},
        ylabel near ticks,
        tick label style={font=\small},
        grid style=dashed,
        ymajorgrids=true,
        small,
        width=0.8\linewidth,
        legend style={at={(0.5,-0.25)},anchor=north, legend cell align={left}},
        legend columns=2,
        yticklabel={+\pgfmathprintnumber{\tick}}
    ]
        \addplot[softred, mark=x] coordinates {
            (1, 0) (3, 4.0) (5, 5.2) (7, 7.6) (9, 7.6) (11, 8.0) (13, 7.8) (14, 9.0)
        };
        \addplot[softblue, mark=x] coordinates {
            (1, 0) (3, 0.2) (5, 2.0) (7, 4.2) (9, 4.6) (11, 4.8) (13, 5.6) (15, 5.4) (17, 5.4) (19, 5.2) (21, 5.8) (23, 6.0)
        };
        \addplot[softred, mark=o] coordinates {
            (1, 0) (3, 0.7) (5, 1.1) (7, 1.8) (9, 1.4) (11, 2.7)
        };
        \addplot[softblue, mark=o] coordinates {
            (1, 0) (3, 1.1) (5, 0.5) (7, 0.7) (9, 1.1) (11, 2.0) (13, 2.0) (15, 1.8) (17, 2.0) (19, 3.4) (21, 4.1) (23, 5.2) (25, 5.0)
        };
        \legend{MATH + 8b, MATH + 70b, GPQA + 8b, GPQA + 70b}
    \end{axis}
\end{tikzpicture}
            \caption{
                The average performance improvement brought by \ourmethod on MATH and GPQA using Llama3.1.
                $\Delta_\texttt{EM}$ denotes the EM improvement compared with the result using the single format.
            }
            \label{fig:format_scale}
        \end{figure}
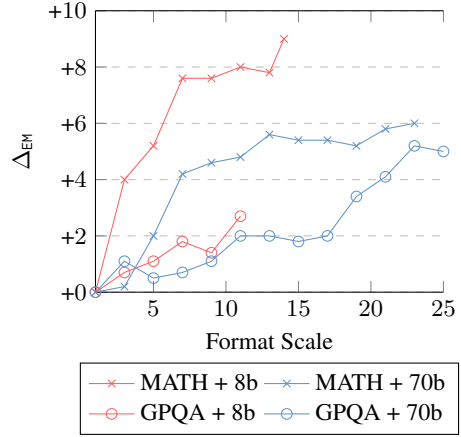

        Considering the computational resource limitations in practical applications, we evaluate the performance of \ourmethod under different scales of reasoning formats.
        The experimental results, as shown in Figure~\ref{fig:format_scale}, reveal that performance consistently improves across different settings as the scale of formats increases, demonstrating the necessity of incorporating more reasoning formats to enhance performance.
        Besides, when a small number ($<5$) of formats are used, \ourmethod also brings a significant improvement, proving the effectiveness of \ourmethod under low computational resources.
        
        Additionally, the performance improvement in each setting follows a trend: it initially increases significantly, then stabilizes, and finally experiences another notable rise.
        This phenomenon can be explained as follows:
        \textit{(i)} Initially, the primary performance bottleneck lies in the inconsistency of LLMs, where increasing the number of reasoning formats enhances the robustness of reasoning, thereby improving the performance.
        \textit{(ii)} Once using a sufficient number of reasoning formats, the performance bottleneck shifts to whether the reasoning formats are suitable for the user question, where adding new formats makes it more likely that the format is suitable for the question, leading to further performance improvements.

    \subsubsection{Format Selection Ratio}
        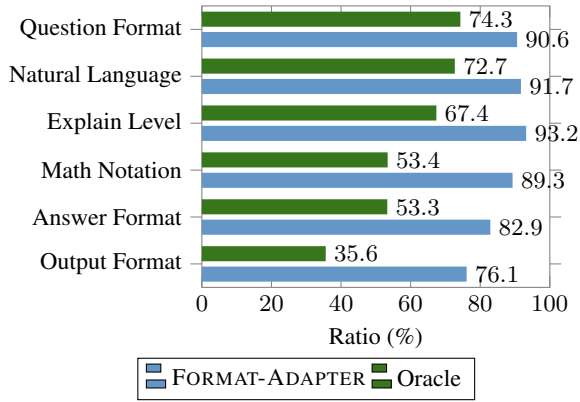
\begin{figure}
            \centering
            \small
            \begin{tikzpicture}
    \begin{axis}[
        xbar,
        xmin=0, xmax=100,
        width=0.8\linewidth,
        xlabel={Ratio (\%)},
        symbolic y coords={
            Output Format, 
            Answer Format,
            Math Notation,
            Explain Level,
            Natural Language,
            Question Format
        },
        ytick=data,
        nodes near coords,
        nodes near coords align={horizontal},
        axis line style={draw=gray},
        bar width=0.2cm,
        legend style={at={(0.3,-0.25)},anchor=north, legend cell align={left}, fill=none},
        legend columns=2,
        enlarge y limits=0.1,
        tick label style={font=\small},
    ]
        \addplot[draw=none, fill=softblue] coordinates {
            (90.6,Question Format)
            (91.7,Natural Language)
            (93.2,Explain Level)
            (89.3,Math Notation)
            (82.9,Answer Format)
            (76.1,Output Format)
        };
        \addlegendentry{\ourmethod}

        \addplot[draw=none, fill=softgreen] coordinates {
            (74.3,Question Format)
            (72.7,Natural Language)
            (67.4,Explain Level)
            (53.4,Math Notation)
            (53.3,Answer Format)
            (35.6,Output Format)
        };
        \addlegendentry{Oracle}
    \end{axis}
\end{tikzpicture}
            \vspace{-1em}
            \caption{
                The average ratio over all datasets of each reasoning format category that is selected by \ourmethod (\textblue{blue}) and that contains the format that can solve the question correctly (Oracle, \textgreen{green}).
            }
            \vspace{-1em}
            \label{fig:format_ratio}
        \end{figure}
    
        To better understand the impact of different reasoning formats on reasoning performance, we compute the ratio of formats selected by \ourmethod or containing the correct answer, as shown in Figure~\ref{fig:format_ratio}. 
        From the figure, we can observe the following:
        \textit{(i)} For different reasoning formats, the ratio selected by \ourmethod follows a trend similar to that of Oracle, indicating that \ourmethod tends to select the appropriate formats, i.e., those that contain the correct answer, thus demonstrating the effectiveness of \ourmethod;
        \textit{(ii)} Compared to the average performance of Oracle with \ourmethod in Table~\ref{tab:main_experiment_single} ($89.4\%$), the best single format still shows a performance gap of $15.1\%$, indicating that different questions suit different formats, suggesting that multiple formats are necessary during reasoning.
        \textit{(iii)} \ourmethod selects a relatively high proportion ($>70\%$) for each category, indicating that LLMs tend to assign higher scores during the Score step, resulting in that \ourmethod selecting many categories that do not contain the correct answer, suggesting the need for further improvement in the scoring method in future work.

    \section{Related Works}
Previous studies have shown that LLMs could exhibit inconsistency during reasoning, producing inconsistent answers when faced with input or parameter perturbations \cite{adiwardana2020humanlikeopendomainchatbot,camburu-etal-2020-make,elazar-etal-2021-measuring}.
To address this issue, \citet{wang2023selfconsistency} proposes Self-Consistency, which generates multiple outputs for the same input and selects the final answer through voting, thereby reducing the impact of perturbations.
Subsequent works have sought to improve upon Self-Consistency to enhance the performance further \cite{li2024escape,besta-etal-2024-graph,wang2024making}. 
For example, Tree-of-Thought~\cite{yao2023tree} decomposes the reasoning process and ensures consistency at each reasoning step as a tree, while DTV~\cite{zhou2024dtv} employs Isabelle formalism to represent answers, improving the accuracy of answer selection.
Notably, many studies have demonstrated that employing diverse reasoning formats to generate answers outperforms relying on a single format to produce multiple outputs \cite{zhang2024flextafenhancingtablereasoning,zhang-etal-2024-autocap,he2024doespromptformattingimpact}. 
For instance, CLSP~\cite{qin-etal-2023-cross} uses different natural language formulations to generate answers, and MultiPoT~\cite{luo2024pythonbestchoiceembracing} leverages multiple programming languages for answer generation.

However, the above methods rely on predefined reasoning formats manually annotated by humans, which can be inefficient and suboptimal, as the most suitable reasoning format varies across questions. 
To address this limitation, we first analyze why utilizing multiple reasoning formats outperforms single-format reasoning and propose an optimization objective based on this insight. 
Guided by this objective, we leverage LLMs to generate and select the most suitable reasoning format, thereby reducing the cost of human annotations and improving reasoning performance.

    \section{Conclusion}
        In this paper, we propose \ourmethod, which generates multiple answers using different reasoning formats, reducing inconsistencies and improving the performance of LLMs. 
        First, we present how to measure reasoning errors when generating multiple answers, showing that multiple reasoning formats outperform a single format. 
        Then, we present \ourmethod, which uses LLMs to generate and select the suitable reasoning formats, improving reasoning performance by reducing the error measurement we present. 
        We conduct experiments on math and commonsense reasoning, where \ourmethod improves performance by an average of $4.3\%$ compared to previous methods, demonstrating its effectiveness.
        We also analyze the relationship between our error measurement and performance, showing a negative correlation that confirms its accuracy in measuring reasoning errors when generating multiple answers.

    \clearpage

    \section*{Limitations}
        \textit{(i)} We have not yet experimented with \ourmethod on more tasks, such as question answering and code generation, where in the future, we will apply \ourmethod to a wider range of tasks to further demonstrate its effectiveness;
        \textit{(ii)} Generating multiple answers incurs significant computational overhead, where in future work, we will explore ways to reduce the computational cost while maintaining or even improving reasoning performance.

    \section*{Ethics Statement}
        All datasets and models used in this paper are publicly available, and our usage follows their licenses and terms.

    \bibliography{anthology}

    \clearpage
    
    \appendix
    \section{Prove of Equations}
    \begin{lemma}
        Let $m, \phi, \bar{\phi}$ follow the definition in \S\ref{subsec:error_single_format}. 
        If $\lim_{m \to \infty} \bar{\phi} = \phi \circ \texttt{f}$, we can derive that $\lim_{m \to \infty} \delta_m = 0$.
        \label{lemma:delta_to_zero}
    \end{lemma}

    \begin{proof}
        Considering that $\bar{\phi} = \texttt{avg}(\phi_i) = \texttt{avg}(\phi \circ \texttt{f} + \delta_i)$, we can derive that 

        $$\texttt{avg}(\phi \circ \texttt{f} + \delta_i) = \phi \circ \texttt{f} (m \to \infty)$$

        Therefore, $\texttt{avg}(\delta_i) = 0 (m \to \infty)$.
        Assume, for contradiction, that \(\lim_{m \to \infty} \delta_m \neq 0\). Then, there exists some \(\epsilon > 0\) such that for large enough \(m\), \(\delta_m \geq \epsilon\).
        For large \(m\), the average of the first \(m\) terms is
        
        \[
        \frac{\delta_1 + \delta_2 + \cdots + \delta_m}{m}
        \]
        
        Since the average tends to 0, for sufficiently large \(m\), we must have
        
        \[
        \frac{\delta_1 + \delta_2 + \cdots + \delta_m}{m} < \epsilon
        \]
        
        However, if infinitely many \(\delta_m \geq \epsilon\), this contradicts the fact that the average tends to 0.
        Thus, \(\lim_{m \to \infty} \delta_m = 0\).
    \end{proof}

    Considering Lemma~\ref{lemma:delta_to_zero}, in the following prove, we substitute $m \to \infty$ with $\delta_m \to 0$.

    \subsection{Prove of Equation~\ref{equ:single_format_error}}
        \label{app:single_error_prove}

        \begin{theorem}
            Let $D, L, m, \phi, \bar{\phi}$ follow the definition in \S\ref{subsec:error_single_format}. We can derive that:
            $$\lim_{\delta_i \to 0} \mathbb{E}_D \left[ L(\bar{\phi}, y) \right] = \frac{1}{m} \sum_{i=1}^{m} L \left( \phi \circ \texttt{f}, y \right)$$
        \end{theorem}

        \begin{proof}
            \begin{align}
                & \frac{1}{m} \sum_{i=1}^{m} \mathbb{E}_D \left[ L \left( \phi_i, y \right) \right] \\
                & = \frac{1}{m} \sum_{i=1}^{m} \mathbb{E}_D \left[ L \left( \phi \circ \texttt{f} + \delta_i, y \right) \right] \\
                & =\mathbb{E}_D \left[ L \left( \phi \circ \texttt{f}, y \right) \right] (\delta_i \to 0)
            \end{align}

            Considering that: 
            
            \begin{align}
                \bar{\phi} & = \frac{1}{m} \sum_{i=1}^{m} \phi_i \\
                & = \frac{1}{m} \sum_{i=1}^{m} \phi \circ \texttt{f} (\delta_i \to 0) \\
                & = \phi \circ \texttt{f}
            \end{align}

            We can derive that:

            \begin{align}
                & \mathbb{E}_D \left[ \frac{1}{m} \sum_{i=1}^{m} L \left( \phi_i, \bar{\phi} \right) \right] \\
                & = \mathbb{E}_D \left[ \frac{1}{m} \sum_{i=1}^{m} L \left( \phi \circ \texttt{f} + \delta_i, \bar{\phi} \right) \right] \\
                & = \mathbb{E}_D \left[ \frac{1}{m} \sum_{i=1}^{m} L \left( \phi \circ \texttt{f}, \phi \circ \texttt{f} \right) \right] (\delta_i \to 0) \\
                & = 0
            \end{align}

            Based on Equation~\ref{equ:ensemble_error}, we can derive that:
    
            \begin{align}
                & \mathbb{E}_D \left[ L(\bar{\phi }, y) \right] \\
                & = \frac{1}{m} \sum_{i=1}^{m} \mathbb{E}_D \left[ L \left( \phi _i, y \right) \right] \\
                & \ \ - \mathbb{E}_D \left[ \frac{1}{m} \sum_{i=1}^{m} L \left( \phi _i, \bar{\phi } \right) \right] \\
                & = \mathbb{E}_D \left[ L \left( \phi \circ \texttt{f}, y \right) \right] (\delta_i \to 0)
            \end{align}
        \end{proof}

    \subsection{Prove of Equation~\ref{equ:multiple_format_error}}
        \label{app:multiple_error_prove}

        \begin{theorem}
            Let $D, L, m, \phi , \bar{\phi }$ follow the definition in \S\ref{subsec:error_multiple_format}.
            we can derive that:
            \begin{align}
                & \lim_{\delta_i \to 0} \mathbb{E}_D \left[ L(\bar{\phi }, y) \right] \\
                & = \frac{1}{m} \sum_{i=1}^m \mathbb{E}_{D} \left[ L(\phi \circ \texttt{f}_i, y) \right] \\
                & \ \ - \mathbb{E}_D[\frac{1}{m} \sum_{i=1}^m L(\phi \circ \texttt{f}_i, \bar{\phi })]
            \end{align}
        \end{theorem}
    
        \begin{proof}
            \begin{align}
                & \frac{1}{m} \sum_{i=1}^{m} \mathbb{E}_D \left[ L \left( \phi _i, y \right) \right] \\
                & = \frac{1}{m} \sum_{i=1}^{m} \mathbb{E}_D \left[ L \left( \phi \circ \texttt{f}_i + \delta_i, y \right) \right] \\
                & = \frac{1}{m} \sum_{i=1}^{m} L \left( \phi \circ \texttt{f}_i, y \right) (\delta_i \to 0)
            \end{align}

            \begin{align}
                & \lim_{\delta_i \to 0} \mathbb{E}_D \left[ \frac{1}{m} \sum_{i=1}^{m} L \left( \phi _i, \bar{\phi } \right) \right] \\
                & = \mathbb{E}_D \left[ \frac{1}{m} \sum_{i=1}^{m} L \left( \phi \circ f_i, \bar{\phi } \right) \right]
            \end{align}

            Based on Equation~\ref{equ:ensemble_error}, we can derive that:
    
            \begin{align}
                & \mathbb{E}_D \left[ L(\bar{\phi }, y) \right] \\
                & = \frac{1}{m} \sum_{i=1}^{m} \mathbb{E}_D \left[ L \left( \phi _i, y \right) \right] \\
                & \ \ - \mathbb{E}_D \left[ \frac{1}{m} \sum_{i=1}^{m} L \left( \phi _i, \bar{\phi } \right) \right] \\
                & = \frac{1}{m} \sum_{i=1}^m \mathbb{E}_{D} \left[ L(\phi \circ \texttt{f}_i, y) \right] \\
                & \ \ - \mathbb{E}_D[\frac{1}{m} \sum_{i=1}^m L(\phi \circ \texttt{f}_i, \bar{\phi })] (\delta_i \to 0)
            \end{align}
        \end{proof}

\section{Prompts of \ourmethod}
    \label{app:prompt}

    \begin{table*}[ht]
        \centering
        \small
        \begin{tabular}{p{0.9\textwidth}}
    \toprule
    \textbf{The prompt of Format Generation} \\
    \midrule
    You are requested to generate possible answer formats that can be changed for the given task, where I want to generate different answers in different formats of the given task. \\
    For each task, you MUST generate the possible answer formats quoted with ** of the task, the number of answer formats of each task MUST > 3. \\
    Here are several examples: \\
    \\
    --- \\
    \\
    Task: Code Generation. \\
    In this task, you are given a question, and then you should generate the Python code to answer the question. \\
    Input:  Today is the last day of the first quarter of 2008. What is the date one year ago from today? \\
    Output: \\
    \texttt{\`}\texttt{\`}\texttt{\`}python \\
    from datetime import datetime, timedelta \\
    today = datetime(2008, 3, 31) \\
    one\_year\_ago = today - timedelta(days=365) \\
    \texttt{\`}\texttt{\`}\texttt{\`} \\
    The possible answer formats that can be changed are: \\
    1. Natural Language: The natural languages of questions can be changed, like change as **Chinese, French, German, Spanish**. \\
    2. Code Language: The code languages of answers can be changed, like change to **Java, C++, R, JavaScript**. \\
    \\
    --- \\
    \\
    Based on the above examples, generate the possible answer formats to be changed for the following task. \\
    \\
    Task: \{task\_name\} \\
    \{task\_definition\} \\
    Output: \{answer\} \\
    \bottomrule
\end{tabular}

\begin{tabular}{p{0.9\textwidth}}
    \toprule
    \textbf{The prompt of Answer Scoring} \\
    \midrule
    Please act as an impartial judge and evaluate the quality of the response provided by an AI assistant to the user question displayed below. Your evaluation should consider correctness and helpfulness. You will be given a assistant's answer. Identify and correct any mistakes. Be as objective as possible. After providing your explanation, you must rate the response on a scale of 1 to 10 by strictly following this format: "$[$rating$]$", for example: "Rating: $[$5$]$". \\
    \\
    $[$Question$]$ \\
    \{question\} \\
    \\
    $[$The Start of Assistant's Answer$]$ \\
    \{answer\} \\
    $[$The End of Assistant's Answer$]$ \\
    \bottomrule
\end{tabular}

        \caption{
            The prompts of \ourmethod.
        }
        \label{tab:prompt}
    \end{table*}

    The prompts of \ourmethod are shown in Table~\ref{tab:prompt}.
    The prompt for the instruction rewriting is provided in the code since this prompt is too long.
    The prompts of the answer generation of each task follow \citet{dubey2024llama3herdmodels}, which can be found in \url{https://huggingface.co/datasets/meta-llama/Llama-3.1-8B-Instruct-evals}.

\section{Baselines of Main Experiments}
    \label{app:baseline}

    \subsection{Single Format}
        \paragraph{Single}
            is to generate one answer using one format with Chain-of-Thought~\cite{wei2022chain}.
            The prompts we used follow \citet{dubey2024llama3herdmodels}.

        \paragraph{Self-Consistency (SC)}
            is similar to Single, while we generate multiple answers for each question.
            The generation number is the same as the format number of \ourmethod for each task and we set temperature as $0.5$, top\_p as $0.9$.
            The prompts are the same with Single.

        \paragraph{Tree-of-Thought (ToT)}
            is to generate the reasoning process step by step, where it votes the results of each step, which is used as the input for the next step.
            The parameters and prompts we used are following the default of the paper.

        \paragraph{DTV}
            asks models to generate Isabelle formulations \cite{nipkow-etal-2002-isabelle} to answer the questions, which can be executed automatically to ensure the logical correctness of the consistent answers.
            The parameters and prompts we used are following the default of the paper.

    \subsection{Multiple Format}
        \paragraph{CLSP}
            asks LLMs to answer the given questions in different natural languages since different questions could suit different languages.
            The natural languages, parameters, and prompts we used follow the default of the paper.

        \paragraph{MultiPoT}
            aims to improve Program-of-Thought~\cite{chen2023program}, which asks LLMs to solve problems with different program languages.
            The program languages, parameters, and prompts we used follow the default of the paper.

        \paragraph{FlexTaF}
            is designed to solve the table reasoning task, which demands LLMs to reason with different tabular formats.
            The table formats, parameters, and prompts we used follow the default of the paper.

\section{Reasoning Formats of \ourmethod}
    \label{app:reasoning_format}

    \begin{table*}[ht]
        \centering
        \tiny
        \begin{tabular}{@{}lcccc@{}}
    \toprule
    \textbf{Model} & \textbf{GSM8K-Hard} & \textbf{MATH500} & \textbf{ARC-C-Hard} & \textbf{GPQA} \\ 
    \midrule
    Llama3.1-8b & 
    \begin{tabular}[c]{@{}l@{}} 
    \textbf{natural language (6)}, code \\ language (2), mathematical \\ notation (2), text format (2), \\ answer style (2), response \\ format (1) 
    \end{tabular} & 
    \begin{tabular}[c]{@{}l@{}} 
    \textbf{natural language (6)}, \\ step-by-step format (3), \\ text format (2), explanation \\ level (1), mathematical \\ notation (2) 
    \end{tabular} & 
    \begin{tabular}[c]{@{}l@{}} 
    \textbf{natural language (8)}, \\ answer format (2), code \\ language (6), explanation \\ level (4), answer style (2), \\ output format (3)
    \end{tabular} & 
    \begin{tabular}[c]{@{}l@{}} 
    natural language (5), \\ answer format (5), \\ \textbf{explanation level (6)}, \\ code language (5), answer \\ style (4), explanation format \\ (5), step-by-step format (4), \\ explanation style (5), \\ mathematical notation (4) 
    \end{tabular} \\ 
    \midrule
    Llama3.1-70b & 
    \begin{tabular}[c]{@{}l@{}} 
    mathematical notation (4), \\ natural language (4), problem \\ format (4), answer format (4), \\ \textbf{reasoning style (3)}, \\ unit of measurement (3) 
    \end{tabular} & 
    \begin{tabular}[c]{@{}l@{}} 
    mathematical notation (3), \\ problem format (5), \\ \textbf{solution approach (3)}, \\ answer format (3), unit \\ system (3), problem \\ complexity (3) 
    \end{tabular} & 
    \begin{tabular}[c]{@{}l@{}} 
    natural language (4), \\ answer format (1), question \\ type (1), answer choice \\ format (4), \\ \textbf{context format (3)}, \\ answer justification (1) 
    \end{tabular} & 
    \begin{tabular}[c]{@{}l@{}} 
    natural language (4), \\ answer format (9), explanation \\ format (1), \\ \textbf{question type (4)}, \\ candidate answer format \\ (7), explanation style (4), \\ answer choice format (7), \\ mathematical notation (6) 
    \end{tabular} \\ 
    \midrule
    GPT-4o & 
    \begin{tabular}[c]{@{}l@{}} 
    natural language (4), \\ mathematical expression (4), \\ \textbf{explanation style (4)}, \\ number representation (5) 
    \end{tabular} & 
    \begin{tabular}[c]{@{}l@{}} 
    \textbf{natural language (6)}, \\ explanation format (2), \\ notation style (2), answer \\ presentation (2), units in \\ solution (2), solution format \\ (3), mathematical \\ representation (3), concluding \\ sentence format (3) 
    \end{tabular} & 
    \begin{tabular}[c]{@{}l@{}} 
    \textbf{natural language (5)}, \\ numerical representation (3), \\ answer structure (2), answer \\ explanation (4), response \\ format (3), question format (4), \\ contextual explanation (2), \\ answer representation (8) 
    \end{tabular} & 
    \begin{tabular}[c]{@{}l@{}} 
    natural language (4), \\ numerical representation (3), \\ answer presentation (2), \\ \textbf{explanation detail (2)}, \\ answer format (3) 
    \end{tabular} \\ 
    \bottomrule
\end{tabular}

        \caption{
            The reasoning categories generated by \ourmethod on different models and datasets.
            The number after each category is the format number corresponding to the category.
            The category with the best performance under each setting is marked in \textbf{bold}.
        }
        \label{tab:format_category}
    \end{table*}

    In this section, we list the reasoning formats generated by different LLMs on various datasets, as shown in Table~\ref{tab:format_category}. 
    We rename some reasoning categories in the experiments of \S\ref{subsec:analysis_experiment} to ensure that the similar categories can be compared together.
    Different formats in Table~\ref{tab:format_category} could be more suitable for different types of questions.
    For instance, for numerical representation, "12" is appropriate for numerical computations, whereas "twelve" is more suitable for non-numerical queries, such as "how many 'e' are in 12?"

    From Table~\ref{tab:format_category}, we can observe that:  
    \textit{(i)} Compared to small-scale LLMs, large-scale LLMs are capable of generating a wider variety of reasoning formats, leading to a more significant performance improvement as demonstrated in Table~\ref{tab:main_experiment_multiple};
    \textit{(ii)} Compared with simple datasets (e.g., GSM8K), a greater number of reasoning formats are generated on more complex datasets (e.g., MATH, GPQA), as more solving approaches are available for complex questions, thus resulting in more diverse reasoning formats.

    Although the number of synthesized formats varies across different tasks and models, the comparability of the results is reliable. 
    That is because, some variation in the number of formats does exist between different tasks; however, these quantitative differences can be considered as variations in the intermediate process, reflecting the model's tendency to synthesize different formats depending on the task. 
    Furthermore, since the experimental setup is consistent across different tasks (e.g., instructions, hyperparameters), we consider the comparability of our results to be reliable.

\section{Efficiency of \ourmethod}
    \label{app:efficiency}

    In this section, we discuss the efficiency of \ourmethod.
    For all experiments conducted on GPT-4o, the total cost is approximately \$600, averaging around \$0.30 per question.
    We focus on two main aspects: the efficiency of the format generation, and the efficiency during inference.

    \subsection{Efficiency during Format Generation}
        Let the number of generated formats be $M$, and $t_\mathcal{M}$ represents the average time that LLM $\mathcal{M}$ takes to process a single data.
        Considering that format generation requires both generation and rewriting, the efficiency of format generation is $2 M t_\mathcal{M} = O(M t_\mathcal{M})$.

        Based on the discussion, we can adjust $M$ to control the efficiency of format generation.
        Furthermore, in practical applications, since format generation is performed offline, the cost of this step can be ignored during online inference.

    \subsection{Efficiency during Reasoning}
        Let the number of formats selected for each query during inference be denoted as $m$, the total number of user queries be $N$, and $t_s$ represents the time to select a single format.
        Since inference involves format selection, answer generation, and answer scoring, the total inference efficiency is given by $N M t_s + 2 m N t_\mathcal{M}$.
        Given that $t_s \ll t_\mathcal{M}$ in practice, the overall inference efficiency simplifies to $O(m N t_\mathcal{M})$.

        It can be observed that the inference efficiency of \ourmethod is comparable to that of Self-Consistency, while \ourmethod offers a significant performance improvement.
        Considering that prior research indicates that there is a positive correlation between model performance and inference time \cite{snell2024scalingllmtesttimecompute,zhong2024evaluationopenaio1opportunities}, it is important to balance efficiency and performance based on the specific application scenario.
        For example, when computational resources are limited, the number of reasoning formats used can be reduced to enhance inference efficiency.

    \subsection{Average Output Tokens of Different Method}
        \label{app:efficiency_output_tokens}
    
        \begin{table*}[ht]
            \centering
            \small
            \begin{tabular}{l|cccc}
    \toprule
    \textbf{Method} & SC & ToT & DTV & \ourmethod \\
    \midrule
    \textbf{Tokens} & $3889.9$ & $24611.4$ & $17816.4$ & $25297.0$ \\
    \bottomrule
\end{tabular}

            \caption{
                The average output tokens per question on MATH using Llama3.1-8b. 
            }
            \label{tab:average_output_token}
        \end{table*}

        To compare the efficiency of \ourmethod with other baselines in practical applications, we measure the average number of tokens output per question, as shown in Table~\ref{tab:average_output_token}. 
        Although \ourmethod is less efficient than Self-Consistency, our method is closer to that of Tree-of-Thought. 
        Considering the performance improvements of \ourmethod over both Self-Consistency and Tree-of-Thought, a balance between efficiency and performance must be considered in practical applications.

\section{Additional Experiments}
    \subsection{Performance Using All Generated Formats}
        \begin{table*}[ht]
            \centering
            \small
            \begin{tabular}{ll|cccc}
    \toprule
    \textbf{Model} & \textbf{Method} & \textbf{GSM8K-Hard} & \textbf{MATH500} & \textbf{ARC-C-Hard} & \textbf{GPQA} \\ 
    \midrule
    \multirow{2}{*}{Llama3.1-8b} & All  & $53.9$ & $54.0$ & $42.2$ & $33.9$ \\ 
                & \ourmethod & $\mathbf{54.7}$ & $\mathbf{56.8}$ & $\mathbf{57.4}$ & $\mathbf{33.9}$ \\ 
    \midrule
    \multirow{2}{*}{Llama3.1-70b} & All  & $73.8$ & $70.2$ & $69.9$ & $47.5$ \\ 
                 & \ourmethod & $\mathbf{76.2}$ & $\mathbf{70.4}$ & $\mathbf{71.5}$ & $\mathbf{51.0}$ \\ 
    \bottomrule
\end{tabular}

            \caption{
                The performance with all formats or the formats selected by \ourmethod.
                All denotes using all generated formats.
                The best performance under each setting is marked in \textbf{bold}.
            }
            \label{tab:all_format}
        \end{table*}
    
        To validate the necessity of the reasoning format selection of \ourmethod, we compare its performance with that of using all formats without selection. 
        The experimental results, as shown in Table~\ref{tab:all_format}, indicate that \ourmethod consistently outperforms that directly using all reasoning formats across all settings, which demonstrates the importance of selecting appropriate reasoning formats.
    
    \subsection{Answer Variability of \ourmethod}
        \begin{table*}[ht]
            \centering
            \small
            \begin{tabular}{ll|cccc}
    \toprule
    \textbf{Model} & \textbf{Method} & \textbf{GSM8K-Hard} & \textbf{MATH500} & \textbf{ARC-C-Hard} & \textbf{GPQA} \\ 
    \midrule
    \multirow{2}{*}{Llama3.1-8b} & Self-Consistency  & $2.0$ & $2.3$ & $1.0$ & $2.3$ \\ 
                & \ourmethod & $\mathbf{12.4}$ & $\mathbf{5.9}$ & $\mathbf{3.6}$ & $\mathbf{3.5}$ \\ 
    \midrule
    \multirow{2}{*}{Llama3.1-70b} & Self-Consistency  & $1.3$ & $1.9$ & $1.0$ & $1.8$ \\ 
                 & \ourmethod & $\mathbf{9.5}$ & $\mathbf{7.8}$ & $\mathbf{2.5}$ & $\mathbf{3.2}$ \\ 
    \bottomrule
\end{tabular}

            \caption{
                The average number of distinct answers generated for each question.
            }
            \label{tab:answer_variety}
        \end{table*}
    
        To demonstrate that \ourmethod can generate different answers using various reasoning formats, we calculate the number of distinct answers generated by our method and Self-Consistency. 
        The results are shown in Table~\ref{tab:answer_variety}, where we observe that, on average, the number of answers generated by our method is significantly higher than that of Self-Consistency. 
        This proves that employing different reasoning formats can guide the model to generate diverse answers.
    
    \subsection{Robustness of \ourmethod}
        \label{app:robustness}
    
        \begin{table*}[ht]
            \centering
            \small
            \begin{tabular}{l|cc|cc}
    \toprule
    \multirow{2}{*}{\textbf{Dataset}} & \multicolumn{2}{c|}{\textbf{8b}} & \multicolumn{2}{c}{\textbf{70b}} \\ 
     & \textbf{Vote} & \textbf{Oracle} & \textbf{Vote} & \textbf{Oracle} \\ 
    \midrule
    GSM8K-Hard & $54.3\pm0.4$ & $89.6\pm0.2$ & $76.1\pm0.3$ & $94.4\pm0.6$ \\ 
    \midrule
    MATH500 & $56.7\pm0.2$ & $74.9\pm0.2$ & $70.2\pm0.4$ & $85.0\pm0.5$ \\ 
    \midrule
    ARC-C-Hard & $57.2\pm0.2$ & $91.2\pm0.4$ & $71.5\pm0.2$ & $88.2\pm0.6$ \\ 
    \midrule
    GPQA & $33.2\pm0.8$ & $93.3\pm0.4$ & $51.1\pm0.7$ & $96.2\pm0.5$ \\ 
    \midrule
    WikiTQ & $55.0\pm0.4$ & $79.3\pm0.5$ & $63.0\pm0.1$ & $83.5\pm0.6$ \\ 
    \bottomrule
\end{tabular}

            \caption{
                The average performance of \ourmethod with five running using Llama3.1.
            }
            \label{tab:robustness}
        \end{table*}
    
        To validate the robustness of \ourmethod, we run it five times with different random seeds, as shown in Table~\ref{tab:robustness}. 
        From the table, it can be observed that the performance of our method does not fluctuate significantly.
        As discussed in \S~\ref{sec:analysis}, generating multiple answers enhances the robustness and reduces performance variability.
    
    \subsection{Evaluation of the Score Quality}
        \begin{table}[ht]
            \centering
            \small
            \begin{tabular}{l|cc}
    \toprule
    \textbf{Dataset} & \textbf{8b} & \textbf{70b} \\
    \midrule
    GSM8K-Hard & $47.7$ & $66.2$ \\
    MATH500 & $46.4$ & $56.0$ \\
    ARC-C-Hard & $52.3$ & $60.2$ \\
    GPQA & $45.7$ & $48.1$ \\
    \bottomrule
\end{tabular}
            \caption{
                The average score quality of \ourmethod with Llama3.1 on different datasets.
            }
            \label{tab:score_quality}
        \end{table}
    
        In this section, we evaluate the quality of the scoring of our method. 
        We use the evaluation metrics $\frac{\sum_{d \in D}{\texttt{Valid}(d)}}{|D|} \times 100$.
        \( D = \{d\} \) represent all the data, and \( \texttt{Valid}(d) \) indicates that if \( d \) is correct, it is represented by the corresponding score; otherwise, it is represented by \( 1 - \) the corresponding score. 
        The statistical results are shown in Table~\ref{tab:score_quality}.
        It can be observed that the evaluation result is not high, which is consistent with the conclusions of previous studies, highlighting the bottleneck of our method. 
        To address this issue, a possible approach is to select the most appropriate reasoning format for each task using the training data first, and then use the selected reasoning format during inference without relying on LLMs for scoring.

    \subsection{Repeated Sampling with Different Formats}
        \begin{figure}
            \centering
            \small
            \begin{tikzpicture}
    \small
    \begin{axis}[
        axis y line*=left,
        xlabel={Sampling Scale},
        xmin=1, xmax=9,
        ymin=47.0, ymax=56.0,
        ylabel={EM},
        ylabel near ticks,
        tick label style={font=\small},
        grid style=dashed,
        ymajorgrids=true,
        small,
        width=0.8\linewidth,
        legend style={at={(0.5,-0.25)},anchor=north, legend cell align={left}},
        legend columns=2,
        yticklabel={+\pgfmathprintnumber{\tick}}
    ]
        \addplot[softblue, mark=x] coordinates {
            (1, 47.8) (3, 50.2) (5, 51.0) (7, 51.6) (9, 51.4)
        };
        \addplot[softred, mark=o] coordinates {
            (1, 47.8) (3, 51.8) (5, 53.0) (7, 55.4) (9, 55.4)
        };
        \legend{Single Format, \ourmethod}
    \end{axis}
\end{tikzpicture}
            \caption{
                The performance of repeat sampling using single format and \ourmethod on MATH with Llama3.1-8b.
            }
            \label{fig:repeat_sampling}
        \end{figure}
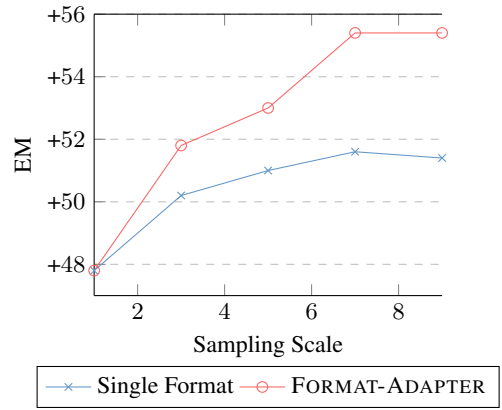
    
        In this section, we evaluate the performance of \ourmethod after repeated sampling, comparing the use of the single format and \ourmethod. 
        The experimental results are shown in Figure~\ref{fig:repeat_sampling}. 
        As can be observed from the figure, when the sampling scale is $> 1$, the performance using a single format is lower than that of \ourmethod, which demonstrates the effectiveness of our method.

\section{Case Study}
    \label{app:case_study}

    \begin{figure}
        \centering
        \includegraphics[width=\linewidth]{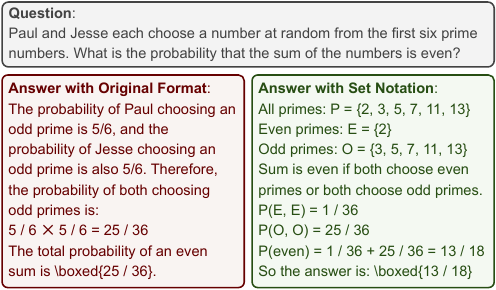}
        \caption{
            An example sampled from MATH answered using different reasoning formats.
            The correct part is marked in \textred{red}, and the incorrect part is marked in \textgreen{green}.
        }
        \label{fig:case_study}
    \end{figure}

    To better understand how \ourmethod improves reasoning performance, we present a case study, as shown in Figure~\ref{fig:case_study}.
    From the figure, it can be observed that when using the original reasoning format, the model overlooks that $2$ is also an odd number, leading to an incorrect answer.
    However, when reasoning with the set notation, the model successfully accounts for all odd numbers, resulting in the correct answer.
    Therefore, utilizing different reasoning formats helps the model approach questions from multiple perspectives and different questions require different reasoning formats. 
    As such, it is essential to integrate various reasoning formats to obtain the correct solution.

\end{document}